\documentclass[conference]{IEEEtran}
\usepackage{cite}
\usepackage{amsmath,amssymb,amsfonts}

\usepackage{algorithmic}
\usepackage{graphicx}
\usepackage{textcomp}
\usepackage{xcolor}
\usepackage{url}

\usepackage{booktabs}
\usepackage{multirow}

\usepackage[utf8]{inputenc} 
\usepackage[T1]{fontenc}    

\usepackage{url}            
\usepackage{booktabs}       
\usepackage{amsfonts}       
\usepackage{nicefrac}       
\usepackage{microtype}      
\usepackage{lipsum}
\usepackage{fancyhdr}       

\usepackage{tabularx}
\usepackage{array}
\usepackage{float}
\usepackage{multirow}
\usepackage{adjustbox}
\usepackage{makecell}
\usepackage{xcolor}
\usepackage{listings}
\usepackage{wrapfig}
\usepackage{booktabs}
\usepackage{caption}
\usepackage{dblfnote}
\usepackage{stfloats}
\usepackage[bottom]{footmisc}
\definecolor{codegray}{gray}{0.95}
\definecolor{stringorange}{rgb}{0.80,0.33,0.0}
\def\BibTeX{{\rm B\kern-.05em{\sc i\kern-.025emb}\kern-.08em
    T\kern-.1667em\lower.7ex\hbox{E}\kern-.125emX}}
    
\usepackage[numbers,sort&compress]{natbib}
\usepackage[
  colorlinks=true,    
  citecolor=green,    
  linkcolor=black,     
  urlcolor=black    
]{hyperref}

\begin{document}

\title{TransportAgents: a multi-agents LLM framework for traffic accident severity prediction\\

}
\author{
  Zhichao Yang\textsuperscript{1}, 
  Jiashu He\textsuperscript{2}
  Jinxuan Fan\textsuperscript{3}
  Cirillo Cinzia\textsuperscript{1}\\[6pt]
  \textsuperscript{1}Civil and Environmental Engineering, University of Maryland\\
  \textsuperscript{2}Computer and Information Science, University of Pennsylvania\\
  \textsuperscript{3}Statistics, University of California, Berkeley\\[6pt]
  $^1\{\text{zcyang97, ccirillo}\}$@umd.edu\\
  $^2$jiashuhe@seas.upenn.edu\\
  $^3$jinxuan\_fan@berkeley.edu
}


\maketitle

\begin{abstract}
    Accurate prediction of traffic crash severity is critical for improving emergency response and public safety planning. 
Although recent large language models (LLMs) exhibit strong reasoning capabilities, their single-agent architectures often struggle with heterogeneous, domain-specific crash data and tend to generate biased or unstable predictions. 
To address these limitations, this paper proposes \textbf{TransportAgent}, a hybrid multi-agent framework that integrates category-specific LLM reasoning with a multilayer perceptron (MLP) integration module. 
Each specialized agent focuses on a particular subset of traffic information—such as demographics, environmental context, or incident details—to produce intermediate severity assessments that are subsequently fused into a unified prediction.

Extensive experiments on two complementary U.S. datasets—the \textit{Consumer Product Safety Risk Management System} (CPSRMS) and the \textit{National Electronic Injury Surveillance System} (NEISS)—demonstrate that TransportAgent consistently outperforms both traditional machine-learning and advanced LLM-based baselines. 
Across three representative backbones, including the black-box commercial models \textit{GPT-3.5-turbo} and \textit{GPT-4o-mini}, and the open-source \textit{LLaMA-3.3-70B-Instruct}, the framework exhibits strong robustness, scalability, and cross-dataset generalizability. 
A supplementary distributional analysis further shows that TransportAgent produces more balanced and well-calibrated severity predictions than standard single-agent LLM approaches, highlighting its interpretability and reliability for safety-critical decision-support applications.
\end{abstract}

\begin{IEEEkeywords}
Traffic Safety, Crash Severity Prediction, Large Language Models (LLMs), Multi-Agent Systems, Hybrid Learning, Interpretability, Machine Learning, Transportation Data Analytics
\end{IEEEkeywords}

\section{Introduction}
Large Language Models (LLMs) \cite{llm1,llm2,llm3,llm4,llm5,llm6} have exhibited the ability to generate coherent narratives, provide answers to diverse queries, and drive autonomous systems across numerous practical domains. These models notably excel in performing general analytical tasks; Recent advances view LLM reasoning as structured planning with world models and benefit from aggregating multiple reasoning trajectories \citep{xiong2025deliberate,xiong2025enhancing}. However, their efficacy diminishes when confronted with tasks that either demand specialized knowledge or involve data with considerable diversity \citep{he2024givestructuredreasoningknowledge, science1,science2,science3,science4}. This variance in performance might be attributed to three specific technical limitations associated with the single-agent architecture of LLMs. Firstly, LLMs frequently do not possess access to the specialized knowledge required to proficiently manage tasks that demand a high degree of expertise; such domain-specific information is infrequently found within the vast resources of Internet data and consequently remains inaccessible during the pre-training phase. This limitation is present irrespective of the model's size, as certain challenges cannot be resolved with just a single forward pass \citep{reasoning_limit}. The absence of this specialized knowledge during the pre-training phase complicates the development of reinforcement learning (RL) post-training powered agents tasked with addressing such complex assignments.

Traffic accident, also known as crash accident or road accident severity prediction, aims to provide timely and accurate prediction of the severity of the accident, usually in terms of fatality or injury levels of the victims. Accuracy of such predictive model is vital to provide insights for the emergency units (policy, medical aid, etc)

In addition, LLM agents tend to exhibit significant biases when dealing with highly variable inputs \citep{biasLLM1,biasLLM2}, which poses significant challenges to tackle tasks in specialized fields such as traffic accident analysis or stock market assessments, where data is inherently diverse and multifaceted, in order to get accurate forecasting, comprehensive reasoning is required from different aspects, it thus remains challenging to develop interpretable models to predict traffic accident severity \citep{Dong2022Predicting}. Furthermore, the creation of reliable k-shot examples for all the distinct facets present in these domains poses a substantial challenge, for human experts, which restricts the deployment of RL-based fine-tuning techniques and prompt-based reasoning methods such as Chain of Thought (CoT) \citep{CoT}, to improve the efficacy of LLM agents.

To address these limitations, this study introduces TransportAgent, a hybrid multi-agent framework that integrates both black-box and open-source LLMs with a structured MLP module for robust and interpretable crash-severity prediction.

\section{Related Work}
Traffic crash severity prediction has been extensively studied, motivated by the need to enhance emergency response, allocate resources, and design preventive countermeasures. 
Over the past decades, the methodological landscape has evolved from interpretable econometric models to advanced machine learning (ML) techniques and, most recently, large language models (LLMs) and hybrid multi-agent frameworks. 
This section reviews the evolution of these approaches in two dimensions: traditional versus LLM-based methods, and general versus traffic-specific applications.

\subsection{Traditional Statistical and Machine Learning Methods}
Traditional econometric models such as binary and multinomial logit, ordered logit, and probit have long been employed to analyze the determinants of injury severity. 
These approaches quantified the impacts of demographic, roadway, and environmental factors on crash outcomes, offering interpretability but limited flexibility~\cite{Xiao2025}. 
Ordered logit/probit models in particular were widely applied to model ordinal injury categories ranging from no injury to fatality~\cite{Zhu2021}. 
Later developments incorporated Bayesian hierarchical methods and random parameters to address unobserved heterogeneity, though these approaches still relied on restrictive linearity assumptions~\cite{AlHashmi2024}.

Machine learning techniques later demonstrated stronger predictive performance by capturing nonlinear interactions and complex dependencies. 
Tree-based methods such as decision trees, random forests, and gradient boosting (e.g., XGBoost) consistently outperformed traditional statistical models across multiple domains~\cite{Islam2022,Hamdan2025}. 
Extensions such as Bayesian-optimized random forests (BO-RF) further enhanced performance and interpretability by fine-tuning hyperparameters systematically~\cite{Yan2025}. 
Neural networks, including multilayer perceptrons (MLPs), were also adopted, and when combined with feature selection, they improved robustness and reduced overfitting. 
More recently, hybrid ML–NLP frameworks that integrate textual features (e.g., NLP-enhanced XGBoost) have shown significant promise in identifying behavioral contributing factors for injury outcomes~\cite{Shao2024}. 
These developments highlight the increasing power of data-driven models to generalize across heterogeneous, high-dimensional inputs.

\subsection{Traditional Methods in Traffic Crash Severity Prediction}
Within the traffic safety domain, the aforementioned methods have been extensively applied to model crash injury severity and related outcomes. 
Econometric frameworks such as ordered logit and probit have historically served as the backbone of crash-severity modeling, quantifying the effects of demographic, roadway, and environmental factors on crash outcomes~\cite{Zhu2021,AlHashmi2024}. 
Subsequent research incorporated random parameters, Bayesian structures, and latent-class specifications to capture unobserved heterogeneity and improve interpretability~\cite{Hamdan2025}. 
Parallel advances in machine learning enabled greater flexibility: random forests, gradient boosting, and neural networks achieved higher accuracy by capturing nonlinear relationships~\cite{Islam2022,Mostafa2025}. 
However, these models often lacked transparency and struggled to incorporate the unstructured narrative data commonly present in crash reports, limiting their reasoning and generalization ability.

\subsection{Large Language Models and General Reasoning Frameworks}
With the advent of large language models (LLMs), researchers began exploring natural-language-based reasoning to extract insights from unstructured text. 
Prompt engineering and Chain-of-Thought (CoT) reasoning~\cite{Wei2023} have been used to guide models such as \textit{GPT-3.5-turbo}, \textit{GPT-4o-mini}, and the open-source \textit{LLaMA-3.3-70B-Instruct} toward better interpretability and step-by-step inference~\cite{Zhen2024,Jaradat2024,llama3.3}. 
Despite these advantages, LLMs exhibit instability and bias when handling heterogeneous data, and their single-agent inference pipelines often lack transparency in intermediate reasoning steps. 
Moreover, the proprietary nature of commercial models restricts reproducibility and integration with structured analytical pipelines. 
Recent efforts in general AI research have introduced hybrid and multi-agent reasoning frameworks, such as MARBLE~\cite{Qasim2025} and GIVE~\cite{he2024givestructuredreasoningknowledge}, which decompose tasks into specialized sub-agents and aggregate their outputs through consensus or supervised learning.

\subsection{LLM-based Methods in Traffic Crash Severity Prediction}
LLM-based frameworks have recently been applied directly to the traffic safety domain. 
Studies such as \textit{CrashSage}~\cite{ZhenYang2025} and \textit{MARBLE}~\cite{Qasim2025} combine structured crash records and textual narratives, using LLaMA-based or GPT-based models to infer injury severity and interpret causal mechanisms. 
Prompt-based LLM frameworks~\cite{Zhen2024,Jaradat2024} further demonstrated that linguistic reasoning can enhance the interpretability of severity prediction. 
Nevertheless, most existing LLM systems rely on single-agent architectures, which limits transparency, domain specialization, and cross-dataset robustness. 

Building upon this trajectory, the proposed \textbf{TransportAgent} framework integrates both black-box and open-source LLMs within a structured multi-agent reasoning pipeline, coupling category-specific reasoning with a supervised MLP integration layer. 
This design achieves interpretable, balanced, and transferable severity predictions that bridge traditional ML interpretability and LLM-based reasoning flexibility.

\section{Transport Agent}
The Transport Agents work closely to embody a hybrid approach that aims to harness the benefits of Large Language Models for understanding inputs from various categories, while also capitalizing on the capabilities of conventional Machine learning techniques to perform supervised, feature learning tasks. It consists of two main components:

(1) \textbf{Multi-Agent Framework}: Every agent possesses a distinct specialization covering a range of disciplines, from data management to severity assessment. Collectively, they form a multi-agent system skilled in optimally utilizing input data to accomplish evaluation objectives.

(2) \textbf{Integration Manager Module}: An integration module, which takes in a vector of intermediary scores from each agent from earlier steps, attempts to learn and adjust these heterogeneous signals to arrive at a final severity evaluation that encapsulates all proposed categories.

\subsection{\textbf{Multi-Agent Framework}}

\begin{figure*}[htbp]
  \centering
  \includegraphics[width=\linewidth]{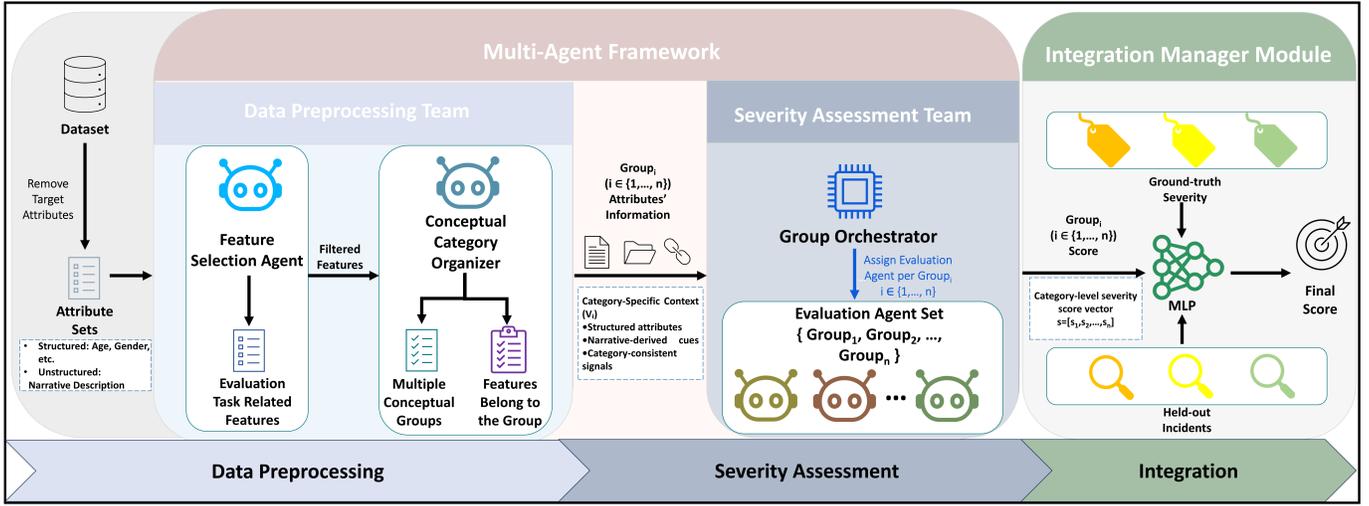}
  \caption{Architecture of the proposed Transport Agent framework. 
  The Data Preprocessing Team selects and organizes relevant features, 
  the Severity Assessment Team produces category-specific evaluations, 
  and the MLP module integrates these into a final severity level score.}
  \label{fig:architecture}
\end{figure*}

To better illustrate the architecture of the proposed Transport Agent,
Figure~\ref{fig:architecture} presents an overview of the complete end-to-end pipeline.

The framework operates as a hybrid multi-agent system that integrates
domain-specific reasoning with supervised machine learning. As shown in
Figure~\ref{fig:architecture}, the architecture consists of three sequential stages:
a Data Preprocessing Team, a Severity Assessment Team, and an
Integration Manager Module.

The Data Preprocessing Team first interfaces with the raw dataset by removing target-related attributes from the input space and separating structured variables (e.g., age, gender) from unstructured narrative descriptions. A Feature Selection Agent then filters attributes based on their relevance to the severity prediction task, ensuring that only task-related information is retained. The selected attributes are subsequently passed to a Conceptual Category Organizer, which assigns features to multiple high-level conceptual groups. Each group contains category-consistent information, combining structured attributes with narrative-derived cues.

The grouped information is then forwarded to the Severity Assessment Team. At the center of this team is a Group Orchestrator, which dynamically assigns a dedicated evaluation agent to each conceptual group. Each agent in the Evaluation Agent Set independently reasons over its assigned group using both structured inputs and contextual narrative signals, producing a category-specific severity score. This parallelized design enables focused assessment of heterogeneous risk factors while maintaining interpretability at the category level.

Finally, the Integration Manager Module consolidates the category-level severity scores into a unified representation. During training, category-specific scores are aligned with ground-truth severity labels. For held-out incidents, the resulting category-level severity score vector is passed to a multilayer perceptron (MLP), which learns to optimally fuse the outputs of the evaluation agents and generate the final severity level prediction. This integration step balances the transparency of agent-based reasoning with the predictive strength of data-driven learning.

The overall workflow depicted in Figure~\ref{fig:architecture} can thus be decomposed into modular components that mirror a hierarchical reasoning process. By separating feature filtering, category-level reasoning, and score integration, the framework ensures that heterogeneous inputs are first standardized and organized before being evaluated in a targeted and interpretable manner. This modular design not only improves robustness and scalability but also allows each specialized agent to focus on a distinct subtask within the severity prediction pipeline.

\begin{figure*}[htbp]
  \centering
  \includegraphics[width=\linewidth]{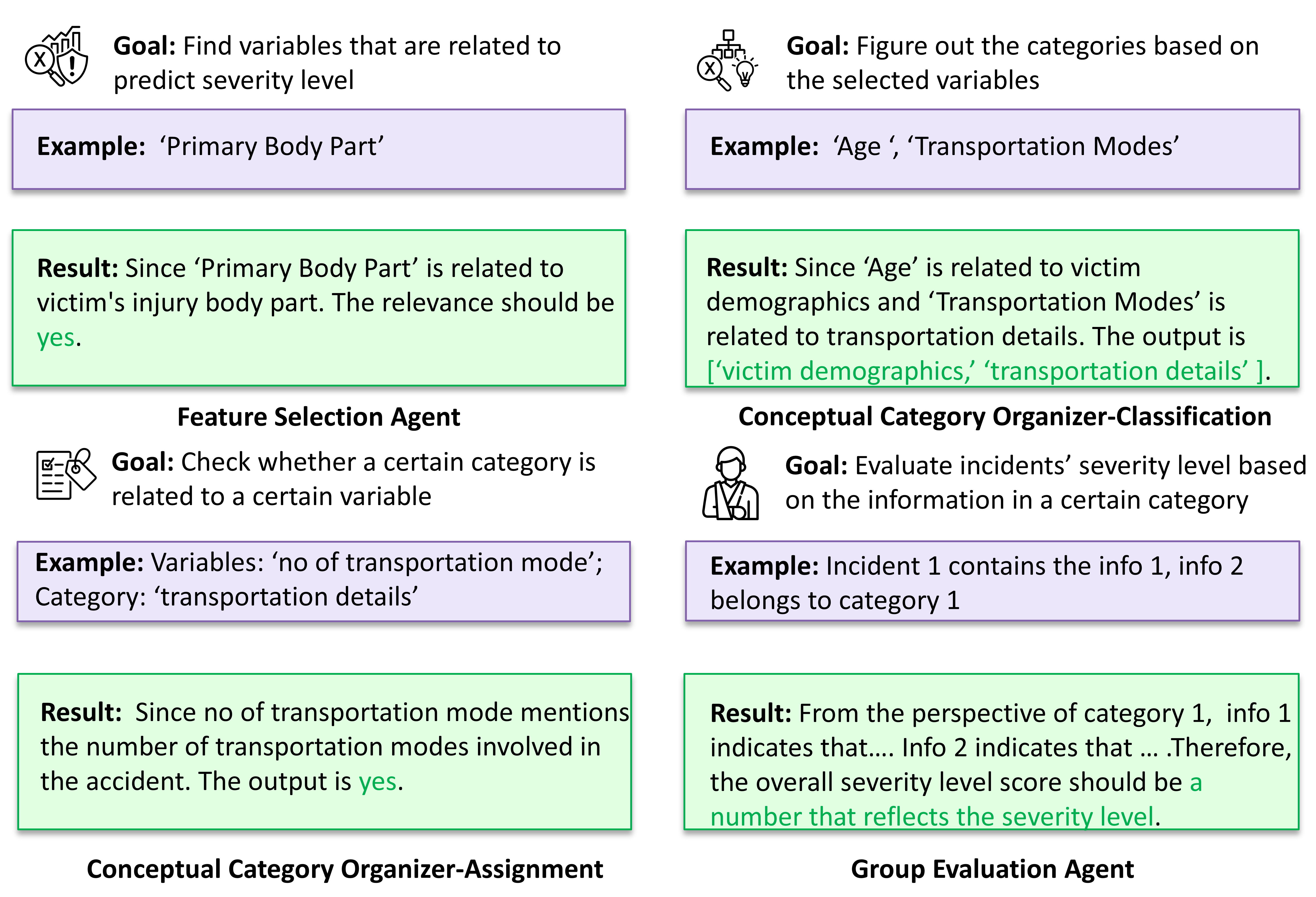}
  \caption{Illustrative roles of Transport Agent components. 
  (Left) \textit{Feature Selection Agent} determines whether a variable is relevant 
  to severity (e.g., \emph{Primary Body Part} $\rightarrow$ relevance = yes). 
  \textit{Severity Prediction Task Linker} checks whether a conceptual category 
  is pertinent to a given variable (e.g., \emph{number of transportation modes} 
  $\rightarrow$ \emph{transportation details}). 
  (Right) \textit{Conceptual Category Organizer} maps selected variables to 
  categories (e.g., \emph{Age} $\rightarrow$ \emph{victim demographics}). 
  \textit{Category Severity Evaluator} reasons over information within one category 
  to output a category-specific severity score.}
  \label{fig:agent-examples}
\end{figure*}

Beyond the high-level pipeline in Figure~\ref{fig:architecture}, we provide a concrete example of agent interactions to clarify the inputs and outputs at each step. Figure~\ref{fig:agent-examples} illustrates how variables are screened for relevance, grouped into conceptual categories, and then evaluated by category-specific agents to produce intermediate severity scores that feed into the integration module.

We now describe 
each team in detail.

\subsubsection{\textbf{Data Processing Team}}

The Data Processing Team is comprised of two specialized agents, whose responsibilities include assisting in the filtration of irrelevant information that could potentially bias goal prediction, and facilitating the classification of diverse attributes.

\textbullet\quad \textbf{Feature Selection Agent:} To enhance clarity, we performed semantic annotation on the variable names along with their associated contents before feeding the candidate features into the agent. For a given set of input attributes, the feature selection agent bases on the evaluation task and its internal expertise to identify features that might affect severity levels. Moreover, it removes unnecessary fields that might cause hallucinations, thereby improving the effectiveness and efficiency of the subsequent tasks\cite{featureselection}. The agent's selected attributes could be characterized as:
\begin{equation}
  \mathcal{V}_{\mathrm{selected}}
  = \mathrm{Agent}\bigl(\hat{\mathcal{V}}, \mathcal{T}\bigr),
\end{equation}
where $\hat{\mathcal{V}}$ is the manually annotated attribute name set, $\mathcal{T}$ denotes the severity level evaluation task.

\textbullet\quad \textbf{Conceptual Category Organizer:} An agent tasked with organizing conceptual categories serves as a preparatory phase by classifying the filtered attributes into multiple conceptual groups, thereby preventing context saturation and preserving focused attention in subsequent tasks. This procedure allows the LLM to mimic human reasoning by segmenting information into manageable parts, evaluating each separately, and subsequently combining these assessments to make a comprehensive judgment, as opposed to processing all information simultaneously.

\subsubsection{\textbf{Severity Assessment Team}}
The Severity Assessment Team consists of several specialized agents, each dedicated to a particular category. Each agent receives information exclusively related to its designated category, which it subsequently analyzes to determine the severity of the incident. The severity level produced by agent $k$ could be defined as:
\begin{equation}
  s_k = \mathrm{Agent}_k\bigl(\mathcal{V}_k,\mathcal{T}\bigr),
\end{equation}
where $\mathcal{V}_k$ is the attributes fall into this category. The advantage is that it could minimize the chance of the model either overlooking or misemphasizing certain details, thereby enabling a more precise focus on particular sub-domains.

\subsection{\textbf{Integration Manager Module}}
Although large language models (LLMs) are highly effective in grasping semantic meaning and identifying patterns, machine learning approaches are superior for efficiently learning numerical relationships and determining weights. We utilized a multilayer perceptron (MLP) to function as the integration manager, which processes the input of per-category scores from the previous step, and consolidates them to produce the ultimate severity prediction output. We split our labeled data into a training set and a testing set, where
\begin{equation}
  \mathcal{D}_{\mathrm{train}}
  = \bigl\{(\mathbf{s}^{(i)},\,y^{(i)})\bigr\}_{i=1}^{N}
\end{equation}
\begin{equation}
  \mathcal{D}_{\mathrm{test}}
  = \bigl\{(\mathbf{s}^{(j)},\,y^{(j)})\bigr\}_{j=1}^{M}
\end{equation}

\subsubsection{\textbf{Training}}
Specifically, we define the input as:
\begin{equation}
  \mathbf{s} = [\,s_{1},\,s_{2},\,\dots,\,s_{d}\,]^\top \in \mathbb{R}^d,
\end{equation}
where each \(s_i\) is the multi-agent generated severity score for category \(i\).  
The MLP then produces a \(C\)-dimensional logit vector, where \(C\) is the number of total severity levels. \(\mathbf{z}^{(i)}\) is the logit vector for the \(i\)-th sample, \(\mathrm{MLP}(\cdot;\theta)\) denotes the multilayer perceptron parameterized by \(\theta\):
\begin{equation}
  \mathbf{z}^{(i)} = \mathrm{MLP}\bigl(\mathbf{s}^{(i)};\mathbf{\theta}\bigr)\;\in\;\mathbb{R}^C,
\end{equation}
Subsequently, we measure how closely those outputs align with the actual labels using the mean cross-entropy loss computed across all ${N}$ training examples\cite{bishop2006pattern}:
\begin{equation}
L(\theta)
= \frac{1}{N} \sum_{i=1}^{N}
  \Bigl[
    -\log\bigl(\mathrm{softmax}\bigl(z^{(i)}\bigr)_{\,y^{(i)}}\bigr)
  \Bigr],
\end{equation}
where $\mathrm{softmax}\bigl(z^{(i)}\bigr)_{\,y^{(i)}}$ is the probability out network assigns to the correct class $y^{(i)}$. Other mispredicted class terms are omitted since $y^{(i)}$ is a one-hot vector.
Finally, the network parameters are updated by taking a gradient step:
\begin{equation}
  \theta \leftarrow \theta - \eta\,\nabla_{\theta}L(\theta).
\end{equation}
where $\eta$ is the learning rate, and $\nabla_{\theta}L(\theta)$ denotes the gradient of the loss $L(\theta)$ with respect to the model parameters $\theta$. This step of gradient descent reduces the cross-entropy loss, thereby enhancing the model's predictions over multiple epochs.

\subsubsection{\textbf{Inference}}At inference time, for each test sample \(\mathbf{s}^{(j)}\in\mathcal{D}_{\mathrm{test}}\) we compute:
\begin{equation}
  \mathbf{z}^{(j)} = \mathrm{MLP}\bigl(\mathbf{s}^{(j)};\theta_{trained}\bigr),
  \quad
  \hat y^{(j)} = \arg\max_{1\le k\le C} \mathbf{z}^{(j)}_k,
\end{equation}
and measure accuracy as
\[
  \mathrm{Acc}
  = \frac{1}{M}\sum_{j=1}^M \mathbf{1}\bigl(\hat y^{(j)} = y^{(j)}\bigr).
\]

\section{Experiments}
\subsection{\textbf{Research Questions}}
This section examines the capabilities of the proposed \textit{TransportAgent} through three guiding 
research questions. Each question naturally connects to different parts of the experimental pipeline,
and the subsequent subsections (B–I) are organized to progressively answer them.

\textbf{RQ1 (Overall Performance).}  
We first ask whether the \textit{TransportAgent} achieves superior severity prediction compared with 
prompting-based LLMs, existing multi-agent systems, and traditional machine-learning models.  
This question motivates the design of the evaluation in Section~E, where all approaches are compared 
on CPSRMS and NEISS. The distributional analysis in Section~F further examines how performance varies 
across severity levels, while Section~B provides the dataset context necessary to interpret the differences.

\textbf{RQ2 (Component Contribution).}  
A second question concerns how the internal components of the framework—feature selection, conceptual 
organization, category-specific agents, and the MLP integrator—interact to produce improved reasoning.  
Insights from Sections~C and~D illustrate the structural properties of the datasets that motivate our 
architectural choices, and Section~I systematically removes components to reveal their individual and 
combined contributions.

\textbf{RQ3 (Robustness and Sensitivity).}  
Finally, we examine which factors influence the stability and reliability of the TransportAgent.  
The distribution-level behavior in Section~G highlights calibration patterns across severity labels, 
while Section~H evaluates robustness under multiple train--test splits.  
The statistical summaries in Sections~C and~D provide additional context for understanding which dataset 
characteristics shape these robustness outcomes.

\subsection{\textbf{Datasets}}
We perform experiments on traffic data from Consumer Product Safety Risk Management System (CPSRMS)\cite{cpsc_chemical_query} and National Electronic Injury Surveillance System (NEISS)\cite{cpsc_neiss_query}. The CPSRMS dataset covers incidents across U.S. from 2017 to 2023. It contains both numerical and narrative-based incident reports submitted by consumers, manufacturers, and health professionals, posing challenges for traditional machine learning models that struggle to represent and process text-based information effectively.  Each record contains descriptive fields such as the incident year, location, and demographic characteristics of those involved (e.g., age and gender), along with a free-text narrative. The NEISS dataset, by contrast, is a structured database that collects data from a statistically representative sample of U.S. hospital emergency departments. It includes product codes that enable clear identification of incident types, along with standardized fields for diagnosis, injury location, patient demographics, and treatment outcomes. For each dataset, we randomly split the data into training and testing sets at a 3:1 ratio. The training set is used to train the multilayer perceptron (MLP) model, while evaluation is performed on the testing set to measure accuracy. 

Each dataset includes an ordinal variable that encodes the severity level of the incident, used as the target label in all experiments.
The definitions for each dataset are summarized in Table~\ref{tab:severity_levels}

\begin{table}[ht]
\centering

\caption{Severity level definitions for \textbf{CPSRMS} and \textbf{NEISS} datasets.}
\label{tab:severity_levels}
\begin{tabular}{@{}c l l@{}}
\toprule
\textbf{Code} & \textbf{CPSRMS} & \textbf{NEISS} \\
\midrule
1 & Incident, No Injury        & Mild \\
2 & Non-admission Medical Care & Moderate \\
3 & Hospital Admission         & Severe \\
4 & Death                      & Fatal \\
\bottomrule
\end{tabular}
\vspace{4pt}
\begin{minipage}{\linewidth}
\end{minipage}
\end{table}
This standardized mapping ensures consistency across datasets and allows direct comparison of model performance under a unified four-level ordinal severity framework.

Both datasets were curated specifically for transportation-related crash incidents, focusing on micromobility (e-bike, bicycle, and other small-vehicle) injuries and their corresponding contextual features. Although CPSRMS and NEISS are general national injury surveillance systems, the subsets used in this study were filtered to include only transportation and roadway crash cases, ensuring consistency with the traffic-safety focus of this research.

\subsection{\textbf{Statistics Description}}

Table~\ref{tab:variables_summary_counts} reports the structured fields shared between the two datasets—
including age group, gender, and severity level—which use standardized categorical definitions and can therefore
be directly compared. As reflected in the table, micromobility-related injuries predominantly involve adult and
male riders, aligning with national transportation injury trends. CPSRMS contains a higher share of severe and
fatal cases due to its consumer-reporting nature, whereas NEISS captures a broader distribution of emergency
department encounters.

Other structured attributes, such as detailed location descriptors or encounter timing, are present in both
datasets but exhibit heterogeneous labels, sparsity, or dataset-specific coding conventions. For example,
location fields mix granular descriptors (e.g., street, home, public facility) with broad or ambiguous entries,
and variables such as road condition or environmental context appear inconsistently across records. To avoid
introducing artificial categories or inconsistent recoding, these fields are not collapsed into unified
representations for the purposes of this summary.

Additional high-value contextual variables—most notably injury causes—lack rigid category boundaries and often
appear in narrative form. Although these cannot be cleanly tabulated without oversimplification, they are fully
utilized in our modeling pipeline through controlled mapping and language-model-based extraction. Thus,
Table~\ref{tab:variables_summary_counts} focuses on structured variables with reliable cross-dataset alignment,
while more granular or unstructured information is incorporated in later stages of feature engineering and
multi-agent reasoning.

\begin{table}[ht]
  \centering
  \caption{Structured Variables: Definitions and Counts}
  \label{tab:variables_summary_counts}
  \setlength{\extrarowheight}{2pt}

  \begin{tabular}{@{}c l c@{}}
    \toprule
    \textbf{Variable} & \textbf{Definition \& Values} & \textbf{Count (CPSRMS / NEISS)} \\
    \midrule

    \multirow{4}{*}{Age}
      & 1: Children ($\leq$14) & 45 / 60 \\
      & 2: Youth (15–24) & 110 / 230 \\
      & 3: Adults (25–64) & 699 / 685 \\
      & 4: Seniors (65+) & 152 / 84 \\
    \midrule

    \multirow{2}{*}{Gender}
      & 1: Female & 186 / 206 \\
      & 2: Male & 961 / 852 \\
    \midrule

    \multirow{2}{*}{Severity Level}
      & 1 & 300 / 500 \\
      & 2 & 180 / 36 \\
      & 3 & 280 / 500 \\
      & 4 & 795 / 23 \\
    \bottomrule
  \end{tabular}
\end{table}

\subsection{\textbf{Initial Explortation}}

\begin{figure}[htbp]
    \centering
    \includegraphics[width=\columnwidth]{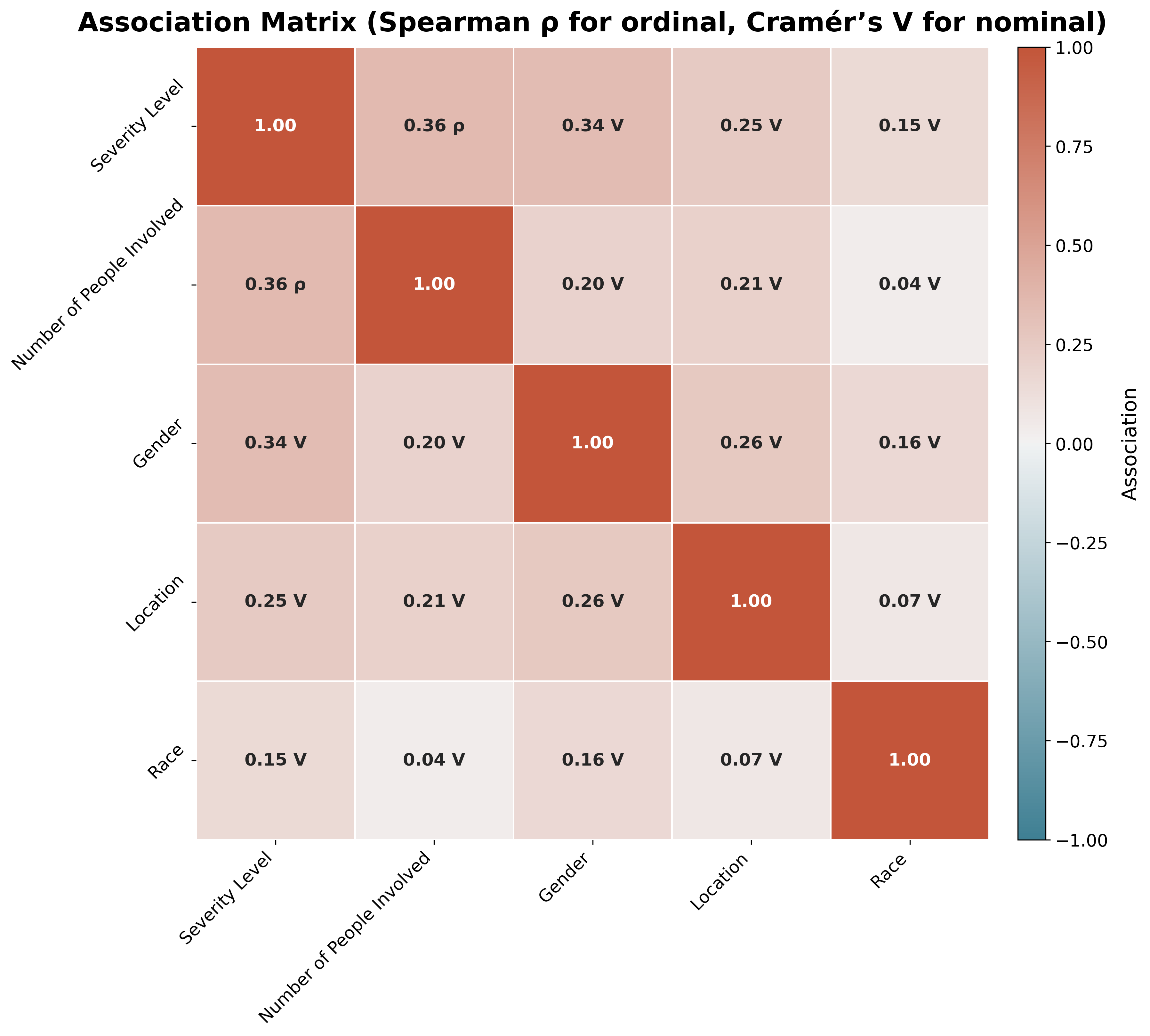}
    \caption{Correlation of variables within category and between severity level in the CPSRMS dataset.}
    \label{fig::cpsrms_correlation}
\end{figure}

\begin{figure}[htbp]
    \centering
    \includegraphics[width=\columnwidth]{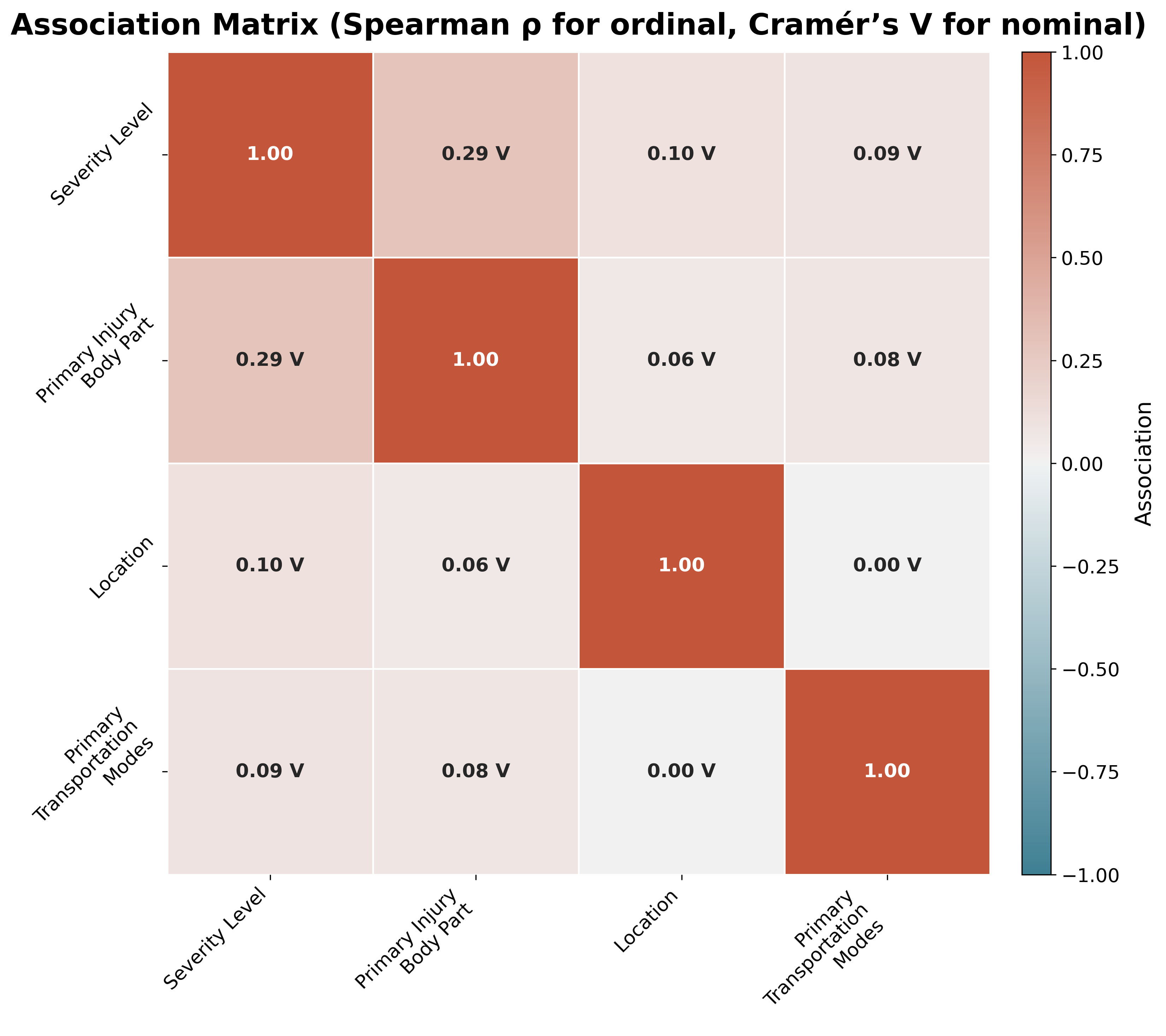}
    \caption{Correlation of variables within category and between severity level in the NEISS dataset.}
    \label{fig::neiss_correlation}
\end{figure}

To examine how variables interact within each conceptual group and how they relate to severity,
we compute a hybrid association matrix using Spearman's $\rho$ for ordinal pairs and
bias–corrected Cramér's $V$ for nominal-variable combinations.
Figures~\ref{fig::cpsrms_correlation} and \ref{fig::neiss_correlation} visualize these associations for CPSRMS
and NEISS respectively.

\textbf{Within–category dependence.}
Across both datasets, variables within the same conceptual category exhibit low interdependence,
indicating minimal multicollinearity and suggesting that each field contributes distinct information.
In \textbf{CPSRMS}, associations among contextual and demographic variables are small to moderate
(e.g., $V\!\le\!0.26$ between \emph{Location} and \emph{Gender};
$V\!\le\!0.21$ between \emph{Location} and \emph{Number of People}).
In \textbf{NEISS}, nominal fields show even weaker connections
($V\!\le\!0.08$ among \emph{Primary Injury Body Part}, \emph{Location}, and
\emph{Primary Transportation Modes}).
These trends validate the architectural choice of assigning distinct category-specific agents, as
the categories largely encode nonredundant signals.

\textbf{Association with severity.}
Both figures also indicate that severity is associated with several structured categories, though the
strength of these relationships is generally modest.  
In \textbf{CPSRMS}, \emph{Severity Level} shows strongest association with 
\emph{Number of People Involved} ($\rho\!\approx\!0.36$) and \emph{Gender} ($V\!\approx\!0.34$),
followed by \emph{Location} ($V\!\approx\!0.25$) and \emph{Race} ($V\!\approx\!0.15$).
In \textbf{NEISS}, severity relates most strongly to \emph{Primary Injury Body Part}
($V\!\approx\!0.29$), with weaker associations for \emph{Location} ($V\!\approx\!0.10$) and
\emph{Primary Transportation Modes} ($V\!\approx\!0.09$).

These findings highlight that although structured variables contain meaningful cues,
their individual associations with severity are limited—reinforcing the need for narrative-derived
context and multi-agent reasoning to capture deeper crash mechanisms.

\subsection{\textbf{Competing baselines and backbone LLMs}}

We conduct a comparative analysis of \textbf{TransportAgent} against both large language model (LLM)–based assessment techniques and conventional machine learning approaches. 
In particular, regarding LLM-based techniques, we incorporated \textit{I/O (k-shot) prompting}~\cite{I/O}, \textit{Chain-of-Thought (CoT) prompting}~\cite{CoT}, and \textit{AutoGen}~\cite{autogen}. 
To ensure a fair comparison, each method was given 10-shot examples formatted according to its specific prompting scheme and evaluated solely on the testing set, so that all methods were tested on identical data partitions.

For machine learning strategies, we employed \textit{multilayer perceptron (MLP)} models both for all numerical features and for a subset of selected features, as well as an \textit{ordered logit (Ologit)} model to represent interpretable econometric approaches traditionally used in injury-severity analysis.

For the backbone LLMs, we evaluate each method using three representative models that collectively span a range of architectures, sizes, and accessibility paradigms: 
the commercial black-box models \textit{GPT-3.5-turbo} and \textit{GPT-4o-mini}, and the open-source \textit{LLaMA-3.3-70B-Instruct}~\cite{llama3.3}. 
This combination covers both proprietary and open ecosystems, allowing us to assess the robustness and adaptability of TransportAgent across distinct LLM families while ensuring reproducibility and transparency in performance evaluation.

\subsection{\textbf{Evaluation Results}}

\textbf{TransportAgent consistently delivers superior severity-prediction performance across datasets, backbone LLMs, and modeling baselines by leveraging structured decomposition and supervised multi-agent fusion.}

Tables~\ref{tab:gpt35_perf}, \ref{tab:gpt4omini_perf}, and \ref{tab:llama33_perf} present comprehensive comparisons of all evaluated approaches on the 
\textbf{CPSRMS} and \textbf{NEISS} datasets across three representative LLM backbones: 
\textit{GPT-3.5-turbo}, \textit{GPT-4o-mini}, and \textit{LLaMA-3.3-70B-Instruct}. 
Because the original NEISS dataset exhibits substantial label imbalance—with extremely sparse representation of certain severity levels—we report results on a balanced NEISS subset. 
This design choice mitigates majority-class dominance and enables a fair and consistent comparison across all methods.

\begin{table*}[ht]
\centering
\caption{Performance of GPT3.5-turbo on CPSRMS, NEISS(Whole), and NEISS(Subset)}
\label{tab:gpt35_perf}
\resizebox{\linewidth}{!}{%
\setlength\tabcolsep{12pt}
\begin{tabular}{lcccccc}
\toprule
\multirow{2}{*}{\# Method} 
  & \multicolumn{2}{c}{CPSRMS} 
  & \multicolumn{2}{c}{NEISS(Whole)} 
  & \multicolumn{2}{c}{NEISS(Subset)} \\
\cmidrule(lr){2-3} \cmidrule(lr){4-5} \cmidrule(lr){6-7}
& Accuracy & Macro-F1 & Accuracy & Macro-F1 & Accuracy & Macro-F1 \\
\textit{Samples} 
& \multicolumn{2}{c}{\textit{1555}}
& \multicolumn{2}{c}{\textit{1059}}
& \multicolumn{2}{c}{\textit{397}} \\
\midrule

\multicolumn{7}{c}{\textbf{Pure LLM-based Reasoning}} \\
1 k-shot (vanilla LLM) 
& 55.31\% & 0.5281 
& 46.7\%  & 0.1679 
& 22.5\%  & 0.3028 \\

2 CoT 
& 62.38\% & 0.5764 
& 41.98\% & 0.3605 
& 36.25\% & 0.4358 \\
\midrule

\multicolumn{7}{c}{\textbf{Augmented LLM-based Reasoning}} \\
3 AutoGen 
& 66.56\% & 0.467 
& 61.85\% & 0.259 
& 37.53\% & 0.285 \\
\midrule

\multicolumn{7}{c}{\textbf{Traditional ML Models}} \\
4 MLP 
& 19.3\% & 0.081 
& 78.7\% & 0.312 
& 63.7\% & 0.46 \\

5 MLP + feature selection 
& 19.3\% & 0.081 
& 79.1\% & 0.277 
& 61.2\% & 0.311 \\
\midrule

\multicolumn{7}{c}{\textbf{Traditional Econometric Model}} \\
6 Ologit 
& 64.6\% & 0.419 
& \textbf{80.4\%} & 0.223 
& 66.2\% & 0.204 \\
\midrule

\multicolumn{7}{c}{\textbf{Hybrid LLM-ML Agent Framework}} \\
7 TransportAgent 
& \textbf{74.6\%}  & \textbf{0.599} 
& 76.41\% & \textbf{0.449} 
& \textbf{67.5\%} & \textbf{0.504} \\
\bottomrule
\end{tabular}
}
\end{table*}

Overall, the results demonstrate the effectiveness, scalability, and robustness of the proposed framework.

First, \textbf{TransportAgent achieves the highest accuracy across all backbones and both datasets}. 
With the open-source \textit{LLaMA-3.3-70B-Instruct} backbone (Table~\ref{tab:llama33_perf}), it reaches 
\textbf{73.31\%} accuracy on CPSRMS and \textbf{76.9\%} on NEISS, outperforming all prompting-based LLM methods and traditional baselines. 
Similarly strong results are observed with smaller commercial models such as \textit{GPT-4o-mini} 
(Table~\ref{tab:gpt4omini_perf}), where TransportAgent achieves \textbf{72.67\%} accuracy on CPSRMS and 
\textbf{75.94\%} on NEISS. These findings indicate that TransportAgent scales reliably across LLMs with different sizes, training paradigms, and accessibility constraints.

Second, the framework performs robustly even in resource-constrained settings. 
Using the lightweight \textit{GPT-3.5-turbo} backbone (Table~\ref{tab:gpt35_perf}), TransportAgent still achieves 
\textbf{74.6\%} accuracy on CPSRMS and \textbf{63.21\%} on NEISS. 
These results confirm that the observed performance gains primarily arise from the architectural design—namely structured task decomposition, category-specific reasoning, and supervised multi-agent fusion—rather than reliance on large or heavily aligned backbone models.

Third, TransportAgent demonstrates stable performance across all evaluated LLM families, including open-source models. 
Across Tables~\ref{tab:gpt35_perf}–\ref{tab:llama33_perf}, TransportAgent consistently outperforms direct prompting strategies and traditional machine-learning approaches, illustrating its ability to enhance LLM reasoning quality without requiring instruction tuning or domain-specific fine-tuning.

Fourth, TransportAgent effectively combines structured attributes with narrative-rich descriptions, enabling richer inference than traditional machine-learning or econometric approaches. 
Its multi-agent semantic processing captures injury mechanisms, environmental context, and latent cues embedded in free-text narratives that are inaccessible to tabular-only models, yielding substantial improvements on both CPSRMS and NEISS.

Fifth, TransportAgent maintains clear performance advantages over general-purpose multi-agent prompting systems. 
Its domain-structured design—integrating specialized category agents with a supervised MLP fusion module—provides greater stability, interpretability, and consistency across model backbones and ecosystems.

Finally, the consistent gains observed across CPSRMS and NEISS underscore the strong generalization ability of TransportAgent. 
The framework learns transferable reasoning patterns that apply robustly to both structured medical records (NEISS) and narrative-heavy consumer reports (CPSRMS), demonstrating adaptability to heterogeneous data modalities and reporting styles.

In summary, TransportAgent establishes a reliable and versatile framework for crash-severity assessment, achieving state-of-the-art performance while preserving interpretability, architectural transparency, and scalability across datasets and backbone LLMs.

\begin{table*}[ht]
\centering
\caption{Performance of GPT4o-mini on CPSRMS, NEISS(Whole), and NEISS(Subset)}
\label{tab:gpt4omini_perf}
\resizebox{\linewidth}{!}{%
\setlength\tabcolsep{12pt}
\begin{tabular}{lcccccc}
\toprule
\multirow{2}{*}{\# Method} 
  & \multicolumn{2}{c}{CPSRMS} 
  & \multicolumn{2}{c}{NEISS(Whole)} 
  & \multicolumn{2}{c}{NEISS(Subset)} \\
\cmidrule(lr){2-3} \cmidrule(lr){4-5} \cmidrule(lr){6-7}
& Accuracy & Macro-F1 & Accuracy & Macro-F1 & Accuracy & Macro-F1 \\
\textit{Samples} 
& \multicolumn{2}{c}{\textit{1555}}
& \multicolumn{2}{c}{\textit{1059}}
& \multicolumn{2}{c}{\textit{397}} \\
\midrule

\multicolumn{7}{c}{\textbf{Pure LLM-based Reasoning}} \\
1 k-shot (vanilla LLM) 
& 64.3\%  & 0.6704 
& 46.7\%  & 0.2506 
& 22.5\%  & 0.3705 \\

2 CoT 
& 65.92\% & 0.6465 
& 41.4\%  & 0.2797 
& 32.5\%  & 0.4298 \\
\midrule

\multicolumn{7}{c}{\textbf{Augmented LLM-based Reasoning}} \\
3 AutoGen 
& 67.5\%  & 0.588 
& 62.04\% & 0.313 
& 34.51\% & 0.354 \\
\midrule

\multicolumn{7}{c}{\textbf{Traditional ML Models}} \\
4 MLP 
& 19.3\% & 0.081 
& 78.7\% & 0.312 
& 63.7\% & 0.46 \\

5 MLP + feature selection 
& 19.3\% & 0.081 
& 79.1\% & 0.277 
& 61.2\% & 0.311 \\
\midrule

\multicolumn{7}{c}{\textbf{Traditional Econometric Model}} \\
6 Ologit 
& 64.6\% & 0.419 
& \textbf{80.4\%} & 0.223 
& 66.2\% & 0.204 \\
\midrule

\multicolumn{7}{c}{\textbf{Hybrid LLM-ML Agent Framework}} \\
7 TransportAgent 
& \textbf{72.67\%} & \textbf{0.686} 
& 74.84\%       & \textbf{0.534} 
& \textbf{71.25\%} & \textbf{0.544} \\
\bottomrule
\end{tabular}
}
\end{table*}
\begin{table*}[ht]
\centering
\caption{Performance of Llama-3.3-70B-Instruct on CPSRMS, NEISS(Whole), and NEISS(Subset)}
\label{tab:llama33_perf}
\resizebox{\linewidth}{!}{%
\setlength\tabcolsep{12pt}
\begin{tabular}{lcccccc}
\toprule
\multirow{2}{*}{\# Method} 
  & \multicolumn{2}{c}{CPSRMS} 
  & \multicolumn{2}{c}{NEISS(Whole)} 
  & \multicolumn{2}{c}{NEISS(Subset)} \\
\cmidrule(lr){2-3} \cmidrule(lr){4-5} \cmidrule(lr){6-7}
& Accuracy & Macro-F1 & Accuracy & Macro-F1 & Accuracy & Macro-F1 \\
\textit{Samples} 
& \multicolumn{2}{c}{\textit{1555}}
& \multicolumn{2}{c}{\textit{1059}}
& \multicolumn{2}{c}{\textit{397}} \\
\midrule

\multicolumn{7}{c}{\textbf{Pure LLM-based Reasoning}} \\
1 k-shot (vanilla LLM) 
& 57.56\% & 0.5225 
& 39.15\% & 0.2497 
& 17.5\%  & 0.3376 \\

2 CoT 
& 63.99\% & 0.6063 
& 32.08\% & 0.3299 
& 25\%    & 0.3896 \\
\midrule


\multicolumn{7}{c}{\textbf{Traditional ML Models}} \\
4 MLP 
& 19.3\% & 0.081 
& 78.7\% & 0.312 
& 63.7\% & 0.46 \\

5 MLP + feature selection 
& 19.3\% & 0.081 
& 79.1\% & 0.277 
& 61.2\% & 0.311 \\
\midrule

\multicolumn{7}{c}{\textbf{Traditional Econometric Model}} \\
6 Ologit 
& 64.6\% & 0.419 
& 80.4\% & 0.223 
& 66.2\% & 0.204 \\
\midrule

\multicolumn{7}{c}{\textbf{Hybrid LLM-ML Agent Framework}} \\
7 TransportAgent 
& \textbf{73.31\%} & \textbf{0.678} 
& \textbf{85.78\%} & \textbf{0.477} 
& \textbf{73.75\%} & \textbf{0.594} \\
\bottomrule
\end{tabular}
}
\end{table*}

\subsection{\textbf{Distributional Comparison of Predicted Severity Levels}}

\textbf{TransportAgent produces more calibrated and class-balanced severity predictions, particularly for higher-severity outcomes that baseline models tend to misrepresent.}

\begin{figure}[!t]
    \centering
    \includegraphics[width=\linewidth]{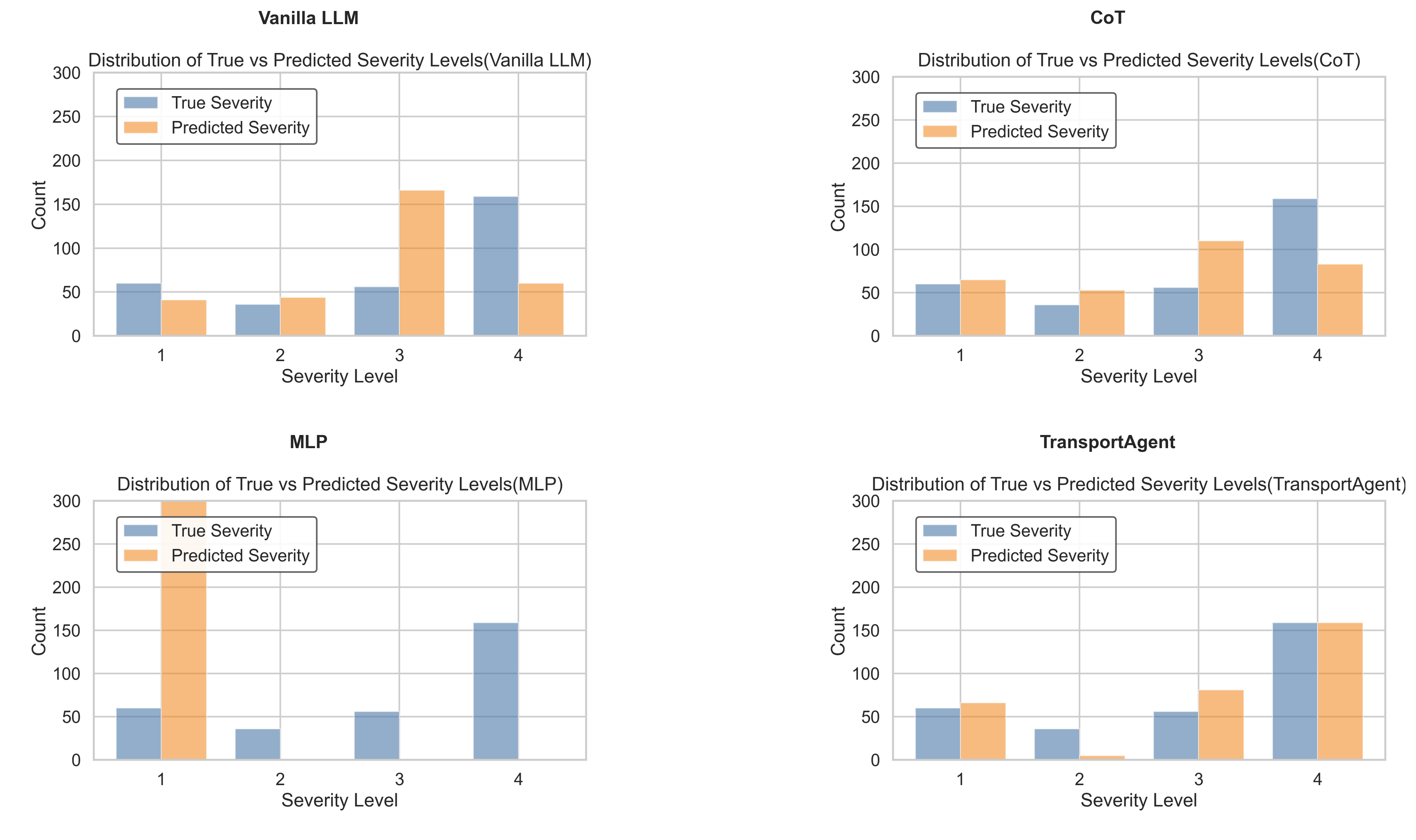}
    \caption{%
        Comparison of true and predicted severity level distributions on the CPSRMS dataset using the GPT-3.5-turbo backbone. Results are shown for Vanilla LLM (top-left), CoT prompting (top-right), MLP (bottom-left), and the proposed TransportAgent framework (bottom-right). Baseline methods exhibit clear distributional bias, particularly under-representation of higher severity levels, whereas TransportAgent produces predictions more closely aligned with the true severity distribution across all levels.}
    \label{fig:severity_dist}
\end{figure}

While Tables~\ref{tab:gpt35_perf}–\ref{tab:llama33_perf} summarize overall accuracy across models and datasets, aggregate metrics alone can obscure systematic biases across individual severity categories. To provide a complementary, distribution-level perspective, Fig.~\ref{fig:severity_dist} compares the distributions of \textit{true} and \textit{predicted} severity levels on the CPSRMS dataset using the GPT-3.5-turbo backbone across four approaches: Vanilla prompting, Chain-of-Thought (CoT) prompting, a standalone MLP, and the proposed TransportAgent framework.

As shown in Fig.~\ref{fig:severity_dist}, both Vanilla and CoT-based LLMs exhibit clear distributional bias, with a tendency to over-predict mid-level severity (Level~3) while under-representing the most severe outcomes (Level~4). Although CoT prompting partially moderates this effect, substantial misalignment with the true severity distribution remains. The standalone MLP demonstrates an even stronger imbalance, collapsing predictions toward the lowest severity class (Level~1) and failing to capture higher-severity cases, reflecting sensitivity to class imbalance and limited semantic context.

In contrast, TransportAgent produces predicted severity distributions that closely track the true class proportions across all four levels. In particular, higher-severity categories are substantially better represented, indicating improved calibration and more stable inference across the severity spectrum. These results suggest that the combination of structured multi-agent reasoning and supervised MLP fusion enables TransportAgent to overcome both prompting-induced bias and purely data-driven imbalance effects.

Overall, the distributional alignment observed in Fig.~\ref{fig:severity_dist} reinforces the quantitative results and demonstrates that TransportAgent improves not only predictive performance but also reliability across severity categories—an essential requirement for downstream traffic safety analysis and decision support.

\subsection{Train--Test Split Robustness Analysis}

\textbf{TransportAgent is robust to variations in train--test partitions, maintaining stable accuracy across repeated random splits and avoiding reliance on a favorable data division.}

Although Table~\ref{tab:gpt35_perf} ~\ref{tab:gpt4omini_perf} ~\ref{tab:llama33_perf} demonstrate consistent improvements across datasets 
and backbones, performance computed on a single train--test split may reflect sampling bias. 
To evaluate whether TransportAgent’s gains generalize beyond any particular partition, 
we conduct a robustness analysis on the CPSRMS dataset using the GPT-3.5-turbo backbone.

Specifically, we repeat the stratified 80/20 split procedure ten times. For each split, only the 
multilayer perceptron (MLP) integration module is retrained, while all LLM-based agents remain fixed. 
This setup isolates variation attributable solely to data partitioning and tests the stability of 
the hybrid architecture under multiple realizations of the dataset.

Table~\ref{tab:robustness_cpsrms} reports the mean and standard deviation of accuracy across the ten runs. 
TransportAgent achieves \textbf{70.69\% $\pm$ 1.76}, outperforming k-shot prompting 
(\textbf{54.04\% $\pm$ 1.08}) and Chain-of-Thought prompting (\textbf{63.22\% $\pm$ 1.97}). 
The notably low variance further confirms that TransportAgent does not depend on a “lucky” split and 
generalizes reliably across different sampled subsets of the data. 

In contrast, direct prompting methods display both lower accuracy and higher variability, highlighting 
the importance of structured multi-agent reasoning and the stabilizing role of supervised fusion.

\begin{table}[ht]
\centering
\caption{Train--test split robustness on CPSRMS with GPT-3.5-turbo. 
Accuracy reported as mean~$\pm$~standard deviation over $K{=}10$ stratified 80/20 splits.}
\label{tab:robustness_cpsrms}
\vspace{2mm}
\begin{tabular}{lcc}
\toprule
Method & \#Splits & Accuracy (\%) \\ 
\midrule
k-shot (vanilla LLM) & 10 & 54.04 $\pm$ 1.08 \\
CoT prompting        & 10 & 63.22 $\pm$ 1.97 \\
\midrule
\textbf{TransportAgent} & 10 & \textbf{70.69 $\pm$ 1.76} \\
\bottomrule
\end{tabular}

\footnotesize{Each split trains a new MLP integration module; LLM backbone (GPT-3.5-turbo) remains fixed.}
\end{table}

\subsection{\textbf{Ablation Study}}

\textbf{TransportAgent benefits from complementary multi-agent components whose removal leads to systematic degradation in both accuracy and macro-F1, demonstrating that its architectural elements operate synergistically rather than independently.}

\begin{figure}[htbp]
    \centering
    \includegraphics[width=\columnwidth]{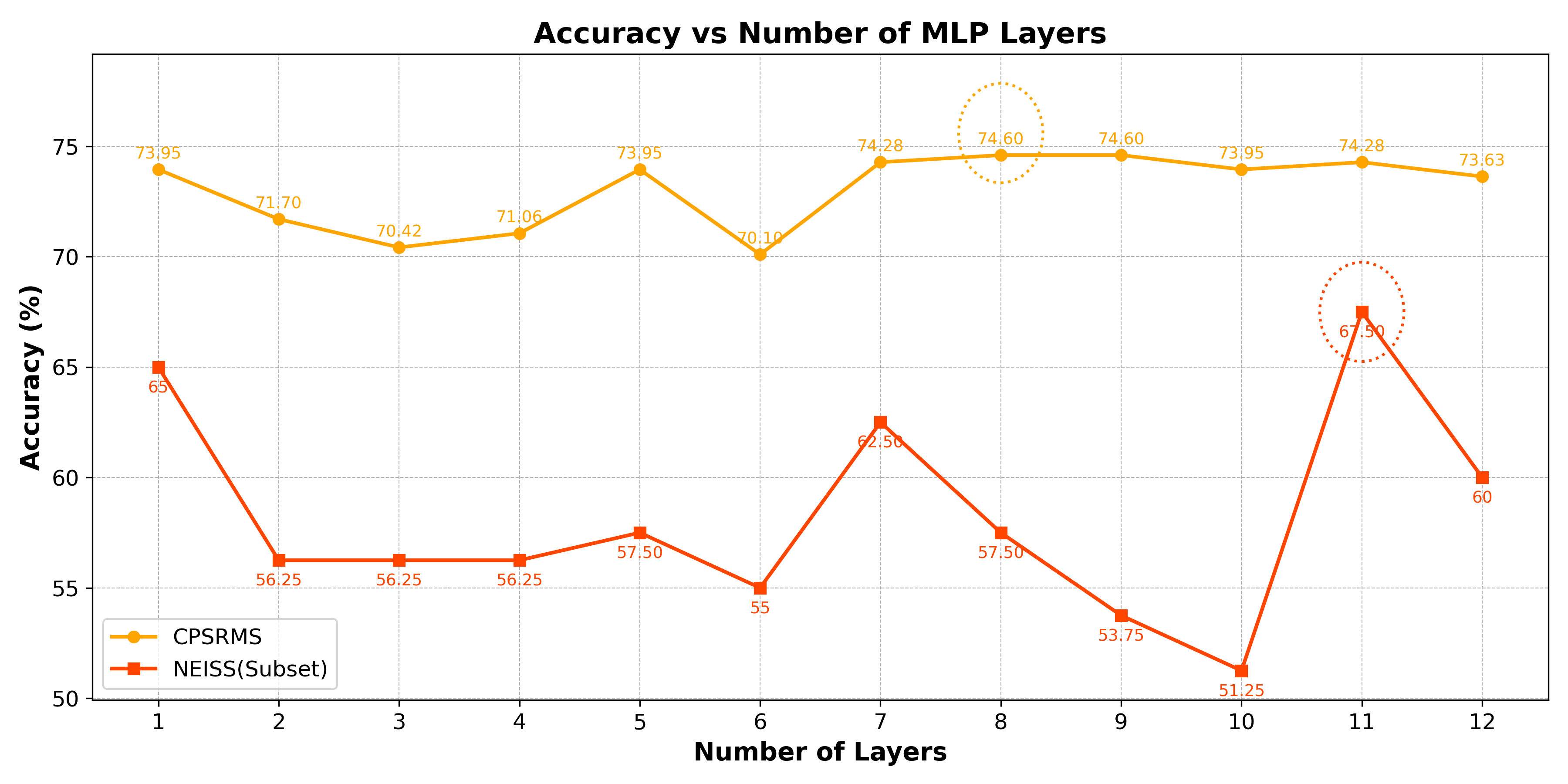}
    \caption{Category-wise accuracy comparison between CPSRMS and NEISS datasets.}
    \label{fig:category_accuracy_comparison}
\end{figure}

\begin{figure}[htbp]
    \centering
    \includegraphics[width=\columnwidth]{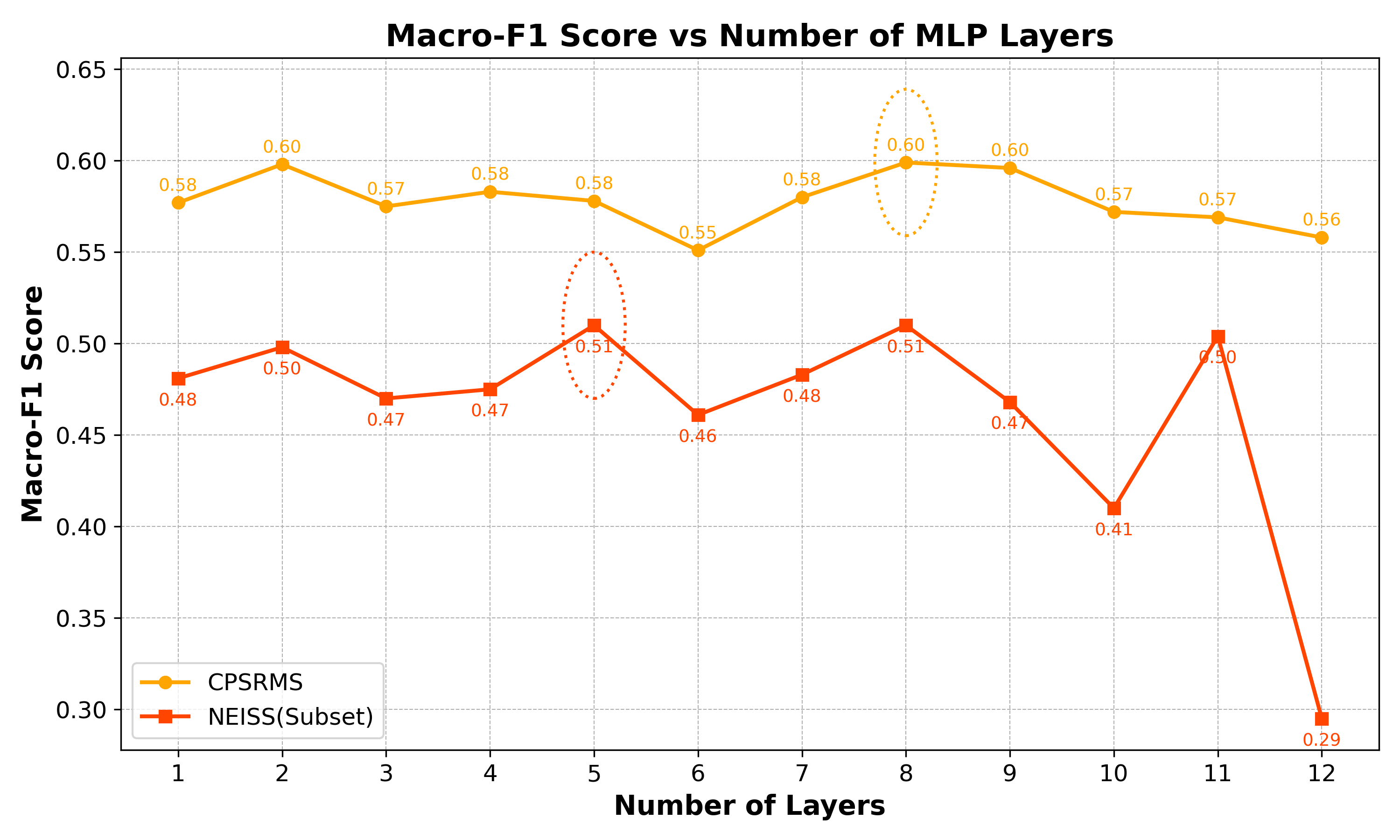}
    \caption{Category-wise macro-f1 score comparison between CPSRMS and NEISS datasets.}
    \label{fig:category_macrof1_comparison}
\end{figure}

\begin{figure}[htbp]
     \centering
     \includegraphics[width=\columnwidth]{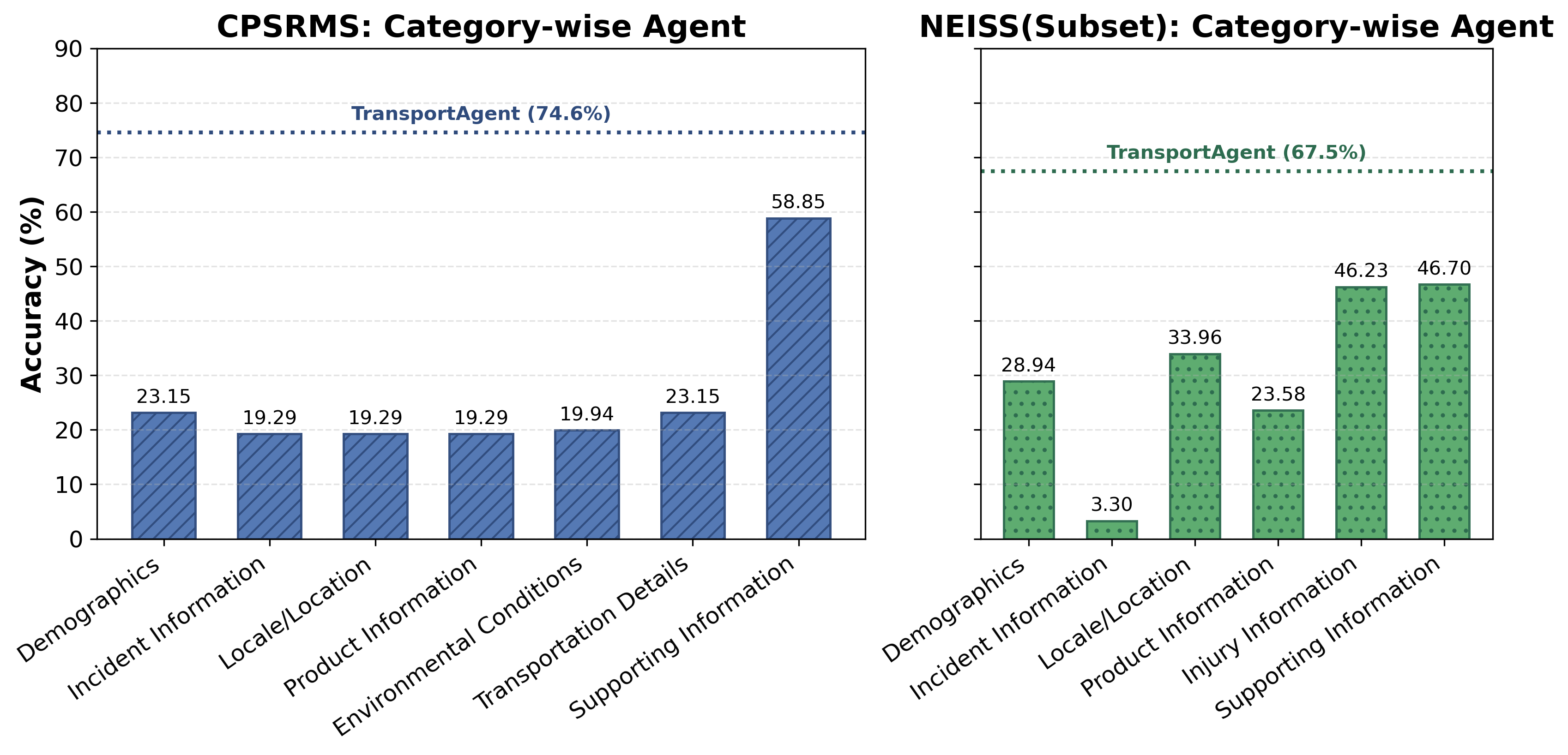}
    \caption{Category-wise Accuracy}
    \label{fig:categoryaccuracy}
\end{figure}

\begin{figure}[htbp]
     \centering
     \includegraphics[width=\columnwidth]{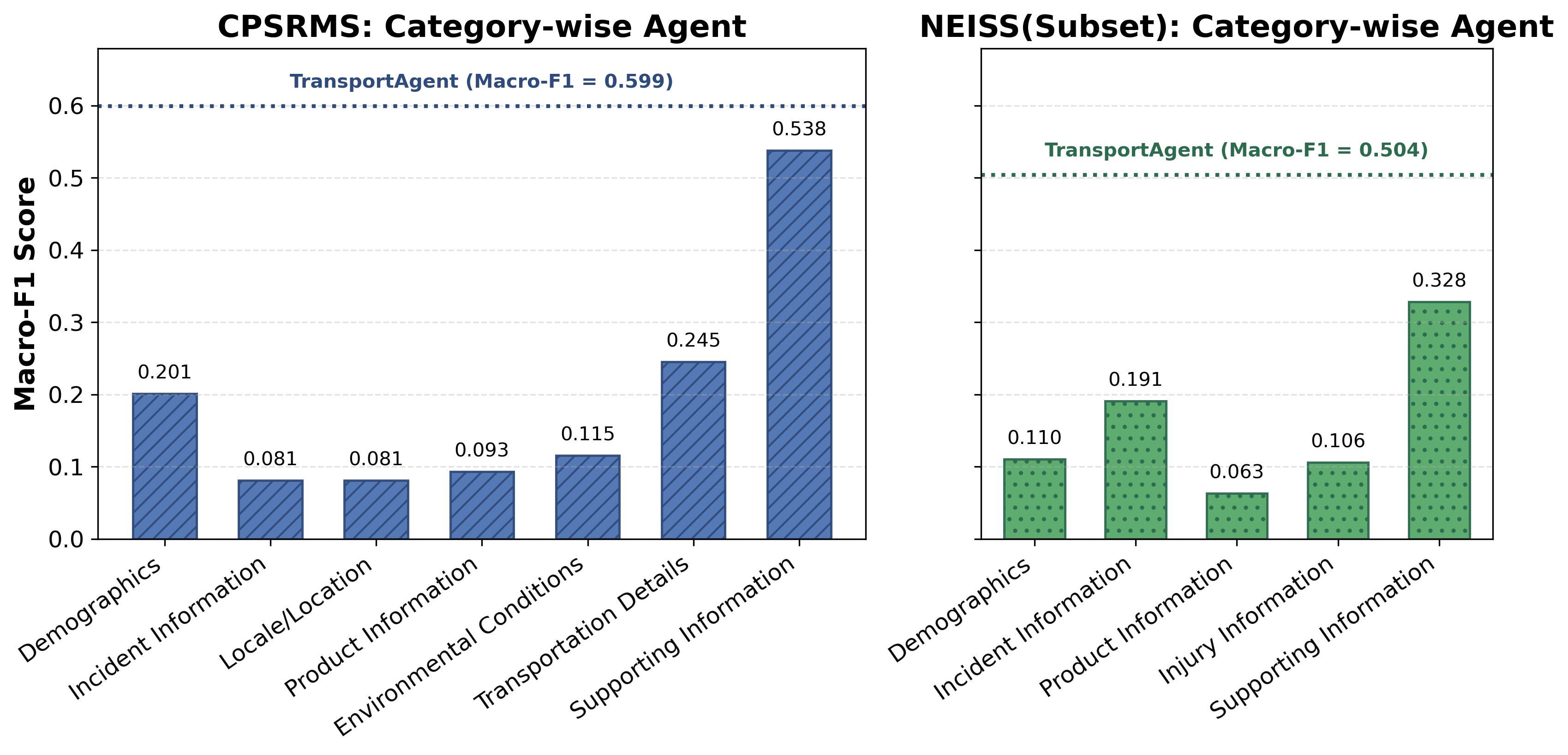}
    \caption{Category-wise Macro-F1 Score}
    \label{fig:categorymacrof1}
\end{figure}

To quantify the contribution of each component within TransportAgent, we conduct a comprehensive ablation study, with results summarized in Figs.~\ref{fig:category_accuracy_comparison}–\ref{fig:categorymacrof1}. 
These analyses examine the effects of removing or isolating key architectural modules—namely feature selection, conceptual category organization, and category-specific severity-assessment agents—using both \textit{accuracy} and \textit{macro-F1} as evaluation metrics on CPSRMS and NEISS.

Figures~\ref{fig:category_accuracy_comparison} and~\ref{fig:category_macrof1_comparison} present category-wise performance comparisons under different architectural variants. 
When either the feature-selection stage or the conceptual-organization stage is excluded, we observe pronounced performance degradation across multiple categories, with particularly large drops in text-dominant domains such as \textit{incident description} and \textit{injury cause}. 
While the accuracy reductions indicate weakened overall predictive capability, the corresponding declines in macro-F1 reveal a more critical insight: removing structured preprocessing disproportionately harms minority and high-severity classes, leading to less balanced and less reliable predictions. 
These results confirm that structured filtering and category organization are essential not only for improving average accuracy, but also for maintaining class-balanced reasoning across heterogeneous crash narratives.

Figures~\ref{fig:categoryaccuracy} and~\ref{fig:categorymacrof1} further isolate the contributions of individual category-specific severity-assessment agents. 
Removing any specialized agent—such as those responsible for \textit{demographics}, \textit{environmental conditions}, or \textit{incident context}—results in a clear performance decline within its corresponding domain. 
Notably, macro-F1 exhibits a larger relative decrease than accuracy in these cases, indicating that the absence of specialized agents particularly impairs the model’s ability to correctly identify less frequent but safety-critical severity outcomes. 
This finding demonstrates that each agent captures distinct, non-redundant knowledge and that their coordinated operation is necessary for stable and balanced severity assessment.

Across all ablation settings, the full TransportAgent consistently achieves the most balanced performance profile, characterized by higher accuracy, improved macro-F1, and reduced inter-category variance relative to partial configurations. 
The MLP-based integration manager plays a crucial role in consolidating these complementary signals, effectively learning how to weight category-specific assessments to produce a unified, calibrated prediction.

Overall, the ablation results validate that TransportAgent’s components function as an integrated system rather than a collection of independent modules. 
Feature selection and conceptual organization ensure high-quality, domain-relevant inputs; category-specific agents provide fine-grained and interpretable reasoning; and supervised MLP fusion reconciles these heterogeneous signals into robust final predictions. 
Their combined effect yields consistently superior and class-balanced performance across both CPSRMS and NEISS, confirming the effectiveness of the proposed hybrid multi-agent framework.

\section{Conclusion}
This paper presented \textbf{TransportAgent}, a hybrid multi-agent framework that unifies the semantic reasoning strength of large language models (LLMs) with the structured learning capacity of conventional machine learning. 
By decomposing traffic incident evaluation into specialized reasoning agents and integrating their intermediate outputs through a multilayer perceptron (MLP), the framework effectively addresses the challenges posed by heterogeneous, narrative-rich, and domain-specific crash data.

Comprehensive experiments on two nationwide datasets, \textit{CPSRMS} and \textit{NEISS}, demonstrate that the proposed framework consistently surpasses both LLM-based prompting strategies and traditional machine learning baselines, including the interpretable econometric \textit{ordered logit (Ologit)} model. 
Across three representative backbones—the commercial black-box models \textit{GPT-3.5-turbo} and \textit{GPT-4o-mini}, and the open-source \textit{LLaMA-3.3-70B-Instruct}—TransportAgent exhibits strong robustness, scalability, and cross-dataset generalizability. 
The framework also produces balanced and well-calibrated severity predictions, mitigating the common tendency of single-agent LLMs to underrepresent high-severity outcomes while preserving interpretability and reasoning transparency.

Each component of TransportAgent contributes meaningfully to its overall performance. 
The feature-selection and conceptual-organization stages ensure domain-relevant, high-quality inputs, while category-specific severity agents capture complementary aspects of crash context that are effectively consolidated by the MLP-based integration manager. 
This synergy enables stable, interpretable, and transferable reasoning across diverse data environments and model architectures.

In summary, TransportAgent demonstrates the potential of multi-agent hybrid reasoning systems for safety-critical applications such as traffic crash severity assessment. 
Beyond transportation safety, the framework offers a generalizable blueprint for combining reasoning-oriented LLMs with structured numerical learning—delivering scalable, transparent, and reliable decision-support solutions applicable to other complex domains such as healthcare, finance, and risk management.

\bibliography{trafficagents}

\end{document}